# Deep learning for *in vitro* prediction of pharmaceutical formulations

Yilong Yang[1,2], Zhuyifan Ye[1], Yan Su[1], Qianqian Zhao[1], Xiaoshan Li[2], Defang Ouyang[1*]

[1]State Key Laboratory of Quality Research in Chinese Medicine, Institute of Chinese Medical Sciences (ICMS), University of Macau, Macau, China

[2]Department of Computer and Information Science, Faculty of Science and Technology, University of Macau, Macau, China

Note: Yilong Yang and Zhuyifan Ye made equal contribution to the manuscript.

Corresponding author: Defang Ouyang, email: defangouyang@umac.mo; telephone: 853-88224514.

## Abstract

Current pharmaceutical formulation development still strongly relies on the traditional trial-and-error approach by individual experiences of pharmaceutical scientists, which is laborious, time-consuming and costly. Recently, deep learning has been widely applied in many challenging domains because of its important capability of automatic feature extraction. The aim of this research is to use deep learning to predict pharmaceutical formulations. In this paper, two different types of dosage forms were chosen as model systems. Evaluation criteria suitable for pharmaceutics were applied to assessing the performance of the models. Moreover, an automatic dataset selection algorithm was developed for selecting the representative data as validation and test datasets. Six machine learning methods were compared with deep learning. The result shows the accuracies of both two deep neural networks were above 80% and higher than other machine learning models, which showed good prediction in pharmaceutical formulations. In summary, deep learning with the automatic data splitting algorithm and the evaluation criteria suitable for pharmaceutical formulation data was firstly developed for the prediction of pharmaceutical formulations. The cross-disciplinary integration of pharmaceutics and artificial intelligence may shift the paradigm of pharmaceutical researches from experience-dependent studies to data-driven methodologies.

[Keywords] pharmaceutical formulation; deep learning; small data; automatic dataset selection algorithm; oral fast disintegrating films; oral sustained release matrix tablets.

## 1. Introduction

The current pharmaceutical industry faces the great pressure of reduced healthcare costs and less number of new active pharmaceutical ingredients (APIs). The pharmaceutical industry should employ more efficient and systematic ways in both drug discovery and development processes.[1] In drug discovery area, scientists now widely use high-throughput screening, combinatorial chemistry, and computer-aided drug design to accelerate drug discovery and development.[2] However, nowadays, pharmaceutical formulation development still strongly relies on the traditional trial-and-error approach by individual experiences of pharmaceutical scientists, which is laborious, time-consuming and cost-expensive. Moreover, it is difficult to achieve the optimum formulations by the trial-and-error studies in the laboratory. Simplification of formulation development becomes essential to formulation scientists. Thus, it is necessary to develop an efficient and systematic method for formulation development to keep pace with the requirements of the pharmaceutical industry.[3]

Machine learning is one of the most exciting research areas in recent years. Machine learning can make data-driven predictions with existed experimental data, which provides a great opportunity for efficient formulation development.[4-9] A well-designed machine learning method can greatly speed the development, optimize formulations, save the cost, keep products consistency, and accumulate and preserve the specific knowledge and expertise from the experts in a well-defined domain.[4] Table 1 summarizes recent progress of machine learning in formulation design. Expert systems (ESs) and artificial neural networks (ANNs) were two useful tools for the formulation development. [4-6, 8, 9] An ES is an intelligent program with the ability to accumulate and preserve the knowledge and experiences of the experts in a specific area (e.g., pharmaceutical formulations).[10] However, it is difficult to extract the vague experiences of pharmaceutical experts into the rules of ESs, and then to accurately predict the performance of a formulation. ANN was the most popular machine learning tool in pharmaceutical formulation prediction. ANN simulates the structure and functions of biological neural networks.[5] ANN is able to solve the problems that are hard to be solved by standard expert systems.[5] However, ANNs still need strong expert knowledge to design feature extractors in the prediction process. In addition, the formulation prediction accuracy by ANNs is relatively low due to the limited experimental data.

**Table 1** Recent progress of machine learning in formulation design

| Machine learning techniques | Formulation | Reference |
| --- | --- | --- |
| Hybrid expert system with ANNs | Hard gelatin capsule formulations | 11 |
| Expert system (SeDeM Diagram) | Orally disintegrating tablets | 12, 13 |
| Expert system with ANNs | Osmotic pump tablets | 5, 14 |
| Ontology-based expert system | Immediate Release Tablets | 15 |
| ME_expert 2.0 | Microemulsions formulations | 6 |
| Fuzzy logic-based expert system | Freeze-dried formulations | 16 |
| Cubist and Random Forest | Cyclodextrin formulations | 17 |

Deep learning as an automatic general-purpose learning procedure has been widely adopted in many domains of science, business, and government.[18] Unlike other machine learning techniques with the requirement of domain expertise to design feature extractors, deep learning can automatically transform low-level representation to more abstract level.[19] Moreover, deep learning is more sensitive to irrelevant and particular minute variations, which make deep learning methods reach higher accuracy rather than other machine learning methods.[20] Deep convolutional networks inspired from visual neuroscience usually achieve good results for processing images, video, speech, and audio.[21, 22] Recurrent neural networks containing history information of the sequence have brought about the breakthrough in sequential data, such as text and speech.[23, 24] Pharmaceutical formulation data include formulation compositions and manufacturing process, which are neither image data nor sequential data. Therefore, the full-connected deep feed-forward network is a good choice for the prediction of pharmaceutical formulations. The recent study showed that deep neural networks (DNNs) made better performance than ANNs with one hidden layer in oral disintegrating tablet prediction.[25] The Maximum Dissimilarity algorithm with the small group filter and representative initial set selection (MD-FIS) selected the representative validation set from the small and imbalanced oral disintegrating tablet data. However, more comparisons of deep learning with other machine learning methods for formulation prediction are still lacked.

In the past five years there are increasing applications of deep learning in pharmaceutical researches.[26-29] The first study of deep learning in pharmaceutical research in 2013 was to compare deep learning with other machine learning approaches by predicting the water solubility of drugs.[30] The results showed that deep learning had better performance than other approaches. After that, more pharmaceutical applications of deep learning were reported. For example, a deep convolution network was developed to predict the epoxidation reactivity of molecules to reduce the drug toxicity.[31] Deep learning was also applied for the prediction of drug-induced liver injury with good performance.[32] Another study also showed that deep learning performed better in toxicity prediction than other computational methods (naive Bayes, support vector machines, and random forests) in 2016 Tox21 Data Challenge. Deep learning was also used in drug discovery.[33-35] DNNs were able to make better predictions than other machine learning approaches on quantitative structure activity relationships (QSAR) data sets.[33] Moreover, multitask deep learning and one-shot learning approaches were used in low data drug discovery, which had better performance than single-task learning.[34, 35] In addition, applying deep learning to mine the increasing datasets in drug discovery not only enables us to learn from the past but to predict future drug repurposing.[36, 37] Recently, the performance of five machine learning models and four DNNs with 2, 3, 4, 5 hidden layers were evaluated on 8 datasets.[38] Further analysis was carried out by using the ranked normalized scores including seven classic measurements of model performance. The final results of the scores ranked by metric and by dataset indicated that the DNNs with five and four hidden layers made better performance than other machine learning approaches. Nearly all reports in recent 5 years suggested that deep learning had more benefits in predictive performance than other machine learning methods.

In this paper, deep learning was applied to pharmaceutical formulation prediction by constructing regression models. One of the main difficulties in formulation prediction is the small dataset with imbalanced input space due to the limited experimental data. For better performance, the data splitting algorithm and the evaluation criteria suitable for pharmaceutical formulation data were introduced. The DNNs were trained on the data of two types of pharmaceutical dosage forms, including oral fast disintegrating films (OFDF) and oral sustained release matrix tablets (SRMT). Comparisons of deep learning with other six machine learning techniques were carried out. Compared with other machine learning methods, deep learning can find out the intricate correlation between pharmaceutical formulations and in vitro characteristics, which shows wide prospects for the application of deep learning in pharmaceutical formulation prediction.

## 2. Methods

*2.1. Pharmaceutical Data*

The pharmaceutical dataset includes 131 formulations of OFDF and 145 formulations of SRMT. The experimental data were extracted from Web of Science. Three different searching terms of oral fast-dissolving films, oral disintegrating films and orodispersible films were used for search literatures about the development of OFDF formulations. The searching strategy of Hydroxypropyl methyl cellulose (HPMC) based sustained release matrix tablet formulations was "HPMC" or "Hydroxypropyl methylcellulose" or "hydroxypropylmethylcellulose" or "hydroxypropylmethyl cellulose" or "hypromellose" and "tablet" or "tablets". The formulation data contain types and contents of both drugs and excipients, process parameters and in vitro characteristics of dosage forms. The characteristics of the two dosage forms were chosen as the prediction targets in this research, including disintegration time for OFDF and cumulative dissolution profiles (2, 4, 6 and 8 hours) for SRMT.

The molecular descriptors were used for representing the properties of APIs. All drugs' name were described with the nine molecular descriptors, including molecular weight, XLogP3, hydrogen bond donor count, hydrogen bond acceptor count, rotatable bond count, topological polar surface area, heavy atom count, complexity and logS. The excipient types were encoded to different numbers. The process parameters include weight, thickness, tensile strength, elongation, folding endurance, actual drug content of OFDF and granulation process, diameter, hardness of SRMT.

*2.2. Data Splitting Strategy*

Three-dataset (training/validation/test datasets) splitting strategy was used in this paper. The training set is for training models, and the validation set is for tuning hyper-parameters to find the best model. The accuracy of the test set shows the prediction ability on unknown data. This strategy is widely adopted in machine learning. For each dosage form, the pharmaceutical data were split into three subsets, both the validation set and the test set include 20 formulations, the rest of the data were used to train the models.

*2.3. Hyperparameters of Machine Learning Methods*

Six machine learning methods were introduced to construct regression models to compare with DNNs, including multiple linear regression (MLR), partial least squared regression (PLSR), support vector machine (SVM), ANNs, random forest (RF) and k-nearest neighbors (k-NN). These regression models were trained using the scikit-learn package.[39] For OFDF, in PLSR, the number of components was set to 8. In ANNs, the networks contained 1 hidden layer with 80 hidden nodes. In RF, the maximum depth of the tree was set to 3. In k-NN, the number of neighbors was set to 5. For SRMT, 4 models were trained simultaneously for 4 time points (2, 4, 6 and 8 hours) by using each machine learning method. The 4 models were developed using the same hyperparameters. In PLSR, the number of components was set to 10. In ANNs, the networks contained 1 hidden layer with 60 hidden nodes. In RF, the maximum depth of the tree was set to 5. In k-NN, the number of neighbors was set to 3.

*2.4. Hyperparameters of Deep Neural Networks*

DeepLearning4j machine learning framework (*https://deeplearning4j.org/*) was used to train the deep neural networks. For OFDF, a feed-forward neural network with 10 layers and 900 epochs was adopted. This network contained 50 hidden nodes on each layer. For SRMT, a feed-forward neural network with 9 layers and 2600 epochs was adopted. This network contained 30 hidden nodes on each layer. All networks chose *tanh* as the activation function except the last layer with *sigmoid* activation function. Learning rate was 0.01. Batch gradient descent with the 0.8 *momentum* was used for training these networks.

*2.5. Evaluation Criteria*

In machine learning, correlation coefficient and coefficient of determination are usually adopted as evaluation metrics for regression problems. Correlation coefficient indicates the linear relationship between two variables. The coefficient of determination shows the correlation between the predicted values and the real values. However, the correlation coefficient and the coefficient of determination cannot well evaluate the performance of the pharmaceutical formulation prediction models. In pharmaceutics the good models for predicting drug dissolution profiles should have less than 10% error [40] Thus, specific criteria suitable for pharmaceutics should be introduced to evaluate the model performance.

Following the FDA (the U.S. Food and Drug Administration) recommendation using the similarity factor *f2* to evaluate the similarity of drug dissolution profiles,[40] the similarity factor *f2* was introduced to evaluate the performance of the models for predicting the cumulative drug release curves. If the *f2* is greater than or equal to 50, it is considered a successful prediction. The accuracy of the cumulative drug release curve prediction is the percentage of the successful predictions in all predictions:

$$accuracy_{CDRC} = \frac{Number(f_2 \geq 50)}{AllPredictions}$$

European Pharmacopoeia stipulates that orodispersible tablets are the tablets that could disperse within 3min (180s). In our dataset, the disintegration time of OFDF ranges from 0 to 100s. Usually, the sucessful prediction is that the error between the predicated time and the experiment time is not higher than 10s. The accuracy of the disintegration time prediction is the percentage of the successful predictions in all predictions:

$$accuracy_{DT} = \frac{Number(|f' - f| \leq 10)}{AllPredictions}$$

Where, $f'$ is the predicted value, $f$ is the experimental value.

## 3. Results and Discussion

Deep learning is a type of representation learning with multiple levels of transform modules, which contains more parameters than other learning algorithms and requires more data for training. However, one of the main difficulties in pharmaceutical formulation prediction is the small dataset with imbalanced input space due to the limited experimental data. Each dosage form has only around 140 formulations. There are 13 APIs in the OFDF dataset, 29 APIs in the SRMT dataset. But near half of the APIs include less than four formulations. Therefore, selecting representative datasets for training and test is very important for the formulation prediction. In our research, the specific evaluation criteria were introduced and several data splitting methods were investigated. Moreover, deep learning was compared with other machine learning techniques for the formulation prediction.

*3.1. Random Data Splitting*

30% data were randomly selected as the validation set, remaining data as the training set. This procedure was repeated 1000 times. However, entirely different accuracy results were obtained, the maximum variation of accuracy was more than 40%, and the average accuracy was less than 60%. In the whole dataset, near half of the APIs include less than four formulations. Therefore, random data splitting algorithm has near 50% probability to select these APIs with less formulations, which may make prediction accuracy quite low and high variation. In short, the random selection algorithm is not suitable for our research and a new approach need to be developed to select the representative data.

### 3.2. Manual Data Splitting

In the manual dataset selection, formulation experts picked up 20 representative data as the validation set for each dosage form. All prediction accuracies on both the training set and the validation set were greater than 90%. However, the manual selection method requires domain knowledge of experts, which is not suitable for large datasets and may have the variation from different experts. Therefore, a selection algorithm should be developed to select the validation set automatically.

### 3.3. Maximum Dissimilarity for Data Splitting

Previous research showed rational selection algorithm can generate better statistical results for the validation set than random selection.[41] Another study indicated that the maximum dissimilarity algorithm was able to select representative test data of compounds from chemical databases.[42] The original maximum dissimilarity algorithm was published in Caret library of R language.[43] In our research, the maximum dissimilarity algorithm was also used to select the validation set. But the test result showed that the maximum dissimilarity algorithm didn't work well on our data, because the accuracies of the validation set were only 83.46% for OFDF and 78.85% for SRMT. After analyzing the splitting result, it was found that the maximum dissimilarity algorithm preferred to select the data from a) the formulations in small API groups, b) boundary formulations, and c) formulations with extraordinary values. The possible reason is that the small API group data, the boundary data or the abnormal data have bigger dissimilar values than other data. Moreover, the maximum dissimilarity algorithm adopts the randomly generated initial set to compute the dissimilarity degree, which is still highly various and not robust due to the small dataset. Therefore, the original maximum dissimilarity algorithm should be improved to select the representative formulation data.

### 3.4. MD-FIS Algorithm for Data Splitting

A new algorithm in R language was developed for selecting the best representative data to validate the models. Figure 1 shows the improved Maximum Dissimilarity algorithm with the small group filter and representative initial set selection (MD-FIS). The MD-FIS algorithm contains 3 steps. In Step 1, the data go through a filter to get rid of the small API group data. In Step 2, the MD-FIS algorithm randomly generates 10000 initial datasets, computes the similarity values between the initial datasets and the remaining datasets, and chooses the initial set with the highest similarity value as the final initial set. In step 3, the final initial set and the remaining data set are used as the input to the dissimilarity algorithm with new cost function. Different from the original cost function, new cost function not only includes the distances (originalDistance) between the candidate data and the initial set, but also contains the distances (subMeanDistance) between the candidate data and the remaining data in the same API group. The new cost function is:

$$\text{cost} = \text{originalDistance} - \alpha * \text{subMeanDistance}$$

Where, $\alpha$ can control the proportion of subMeanDistance, the maximum dissimilarity algorithm selects the data with the maximum cost. The new cost function will prevent the selection of the boundary data. The result was much better than that of the original maximum dissimilarity algorithm. The prediction accuracies were 95.57% for OFDF and 82.02% for SRMT on the validation set.

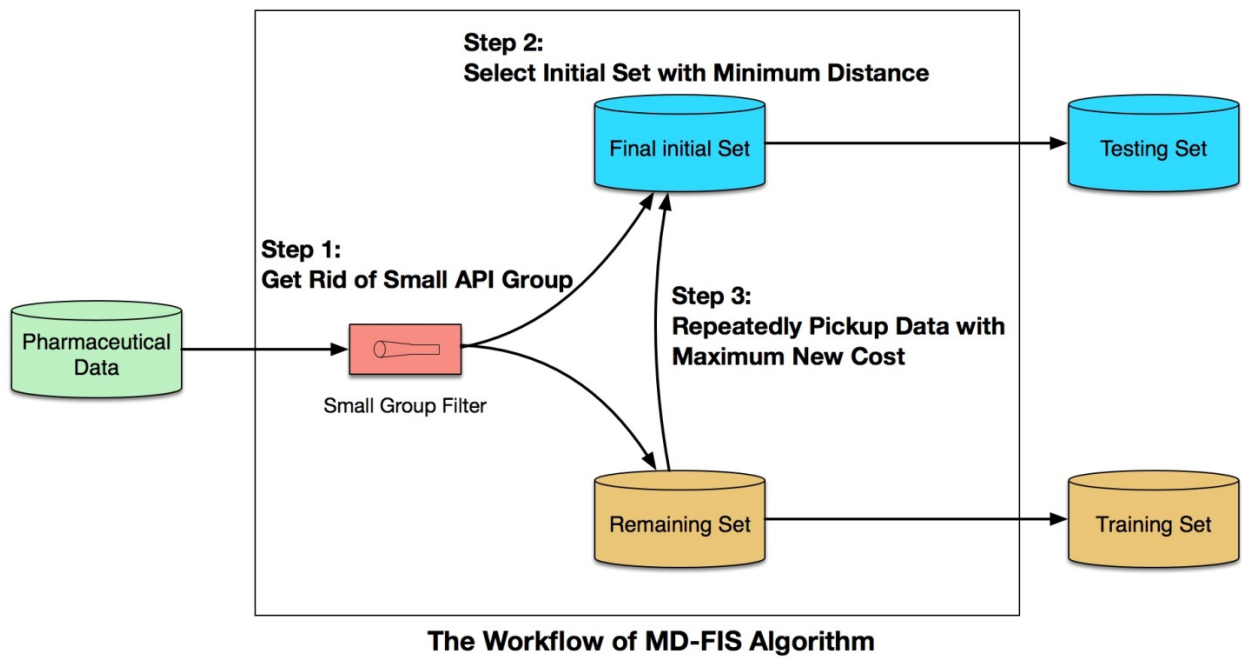

Figure 1 The workflow of MD-FIS algorithm

## 3.5. Comparison of Deep Learning and Conventional Machine Learning Methods

**Table 2** Results of the conventional machine learning models and the deep neural network on the OFDF training, validation and test sets

| Machine learning techniques | Training Set | | | Validation Set | | | Test Set | | |
|---|---|---|---|---|---|---|---|---|---|
| | Accuracy (%) | RMSE | MAE | Accuracy (%) | RMSE | MAE | Accuracy (%) | RMSE | MAE |
| MLR | 90.11 | 0.0671 | 0.0508 | 60.00 | 0.1311 | 0.0999 | 65.00 | 0.1778 | 0.1183 |
| PLSR | 76.92 | 0.0917 | 0.0705 | 70.00 | 0.1136 | 0.0835 | 70.00 | 0.0970 | 0.0705 |
| SVM | 79.12 | 0.1136 | 0.0711 | 70.00 | 0.1308 | 0.0959 | 75.00 | 0.1039 | 0.0795 |
| ANN | 74.73 | 0.1140 | 0.0809 | 70.00 | 0.1105 | 0.0846 | 70.00 | 0.0959 | 0.0772 |
| RF | 84.62 | 0.0775 | 0.0567 | 80.00 | 0.0917 | 0.0721 | 70.00 | 0.1068 | 0.0774 |
| k-NN | 80.22 | 0.0975 | 0.0649 | 75.00 | 0.1025 | 0.0727 | 75.00 | 0.0877 | 0.0608 |
| DNN | 97.80 | 0.0420 | 0.0307 | 80.00 | 0.0842 | 0.0705 | 80.00 | 0.0714 | 0.0565 |

**Table 3** Results of the conventional machine learning models and the deep neural network on the SRMT training, validation, and test sets

| Machine learning techniques | Training Set | | | Validation Set | | | Test Set | | |
|---|---|---|---|---|---|---|---|---|---|
| | Accuracy (%) | RMSE | MAE | Accuracy (%) | RMSE | MAE | Accuracy (%) | RMSE | MAE |
| MLR | 52.38 | 0.1356 | 0.1031 | 35.00 | 0.1212 | 0.1042 | 25.00 | 0.2182 | 0.1685 |
| PLSR | 55.24 | 0.1446 | 0.1066 | 55.00 | 0.1175 | 0.0961 | 45.00 | 0.1609 | 0.1203 |
| SVM | 60.95 | 0.1568 | 0.1013 | 50.00 | 0.1170 | 0.0960 | 45.00 | 0.1559 | 0.1147 |
| ANN | 57.14 | 0.1330 | 0.0998 | 50.00 | 0.1389 | 0.1137 | 50.00 | 0.1497 | 0.1124 |
| RF | 76.19 | 0.0975 | 0.0692 | 55.00 | 0.1308 | 0.1045 | 55.00 | 0.1170 | 0.0908 |
| k-NN | 64.76 | 0.1229 | 0.0825 | 45.00 | 0.1526 | 0.1264 | 40.00 | 0.1565 | 0.1306 |
| DNN | 99.05 | 0.0335 | 0.0237 | 80.00 | 0.0967 | 0.0660 | 80.00 | 0.0902 | 0.0673 |

In this study, the models of MLR, PLSR, SVM, ANNs, RF, k-NN and deep learning were developed on the formulation data. Here, three datasets were split by using the MD-FIS algorithm twice without the need of personal expertise. In the prediction of SRMT, 4 models were built for the 4 time points (2, 4, 6 and 8 hours) by using each of the machine learning methods. The final results of accuracies, root mean squared errors (RMSE) and mean absolute errors (MAE) were shown in Table 2 and Table 3. In the prediction of OFDF, for all the models based on the linear or nonlinear conventional machine learning methods, the accuracies only reach around 70% on the OFDF validation and test sets. As to the MLR model, the accuracies are relatively low than other conventional machine learning models on the OFDF validation and test sets. In the prediction of SRMT, the conventional machine learning models made predictions with the accuracies ranging from 25% to 55% on the SRMT validation and test sets, which are far from the satisfied prediction for the formulation development. In summary, all these six conventional machine learning methods could not achieve enough performance for the OFDF and SRMT formulation prediction.

Here, the training, validation and test sets for training the DNNs are the same as the datasets used for training the previous machine learning models. A multi-label model was built for the 4 time points (2, 4, 6 and 8 hours) using deep learning techniques. As shown in Table 2&3, all prediction accuracies of the deep neural networks were over 80%, which could satisfy the requirements of the formulation prediction. In both OFDF and SRMT predictions, deep learning got the highest accuracies on the training, validation and test sets. Deep learning surpassed other conventional machine learning methods because deep learning including multiple hidden layers could transform the low level representation to higher level features without artificial feature engineering. In SRMT prediction, huge performance improvements of deep learning were found than other machine learning methods. The result indicates that deep learning can greatly improve the model accuracy in multi-label formulation prediction, because deep learning can leverage the shared information among the multiple tasks. Figure 2 shows the experimental and the deep learning predicted disintegration time of the formulations in the OFDF test set. Table 4 lists the $f2$ values between the experimental and the deep learning predicted cumulative drug released curves of the formulations in the SRMT test set. From these figures and tables, it is quite clear that the prediction performance of deep learning is satisfied.

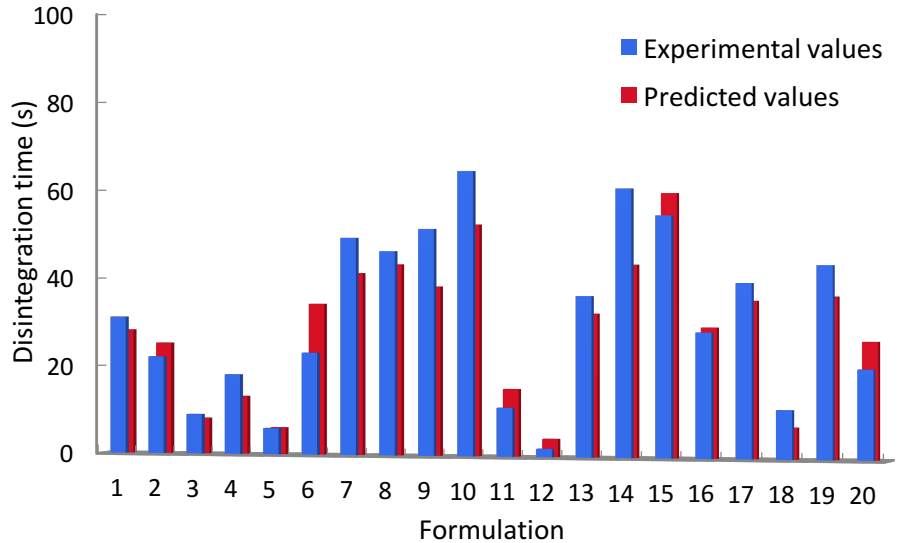

**Figure 2** Comparing the experimental and the deep learning predicted disintegration time of the formulations in the OFDF test set

**Table 4** *f2* values between the experimental and the deep learning predicted cumulative drug released curves of the formulations in the SRMT test set

| Formulation | *f2* values | Formulation | *f2* values |
|---|---|---|---|
| 1 | 77.42 | 11 | 65.72 |
| 2 | 63.35 | 12 | 90.05 |
| 3 | 64.84 | 13 | 57.05 |
| 4 | 67.21 | 14 | 41.91 |
| 5 | 59.75 | 15 | 55.06 |
| 6 | 50.85 | 16 | 65.84 |
| 7 | 77.77 | 17 | 51.08 |
| 8 | 30.39 | 18 | 49.57 |
| 9 | 44.56 | 19 | 64.42 |
| 10 | 74.47 | 20 | 59.35 |

Figure 3, 4, 5 and 6 show the relationship between the experimental and the deep learning predicted results on the OFDF and SRMT training, validation and test sets. It could be seen from these figures that the experimental and the deep learning predicted values are very closed.

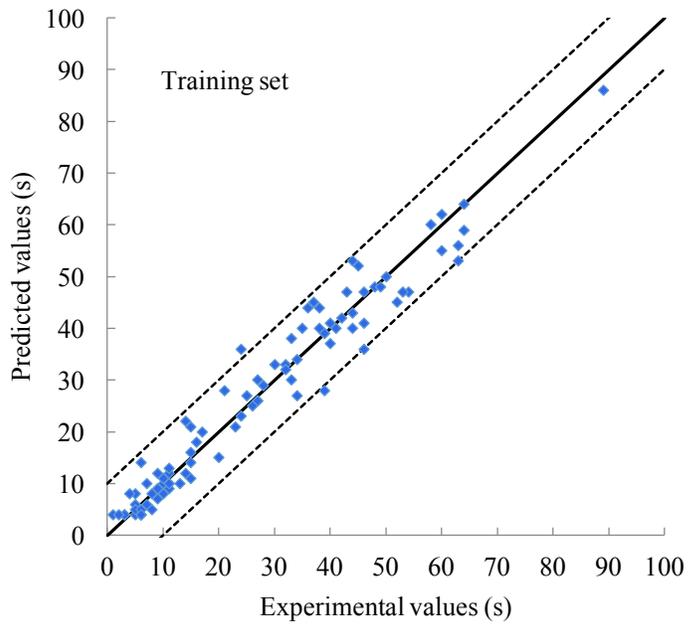
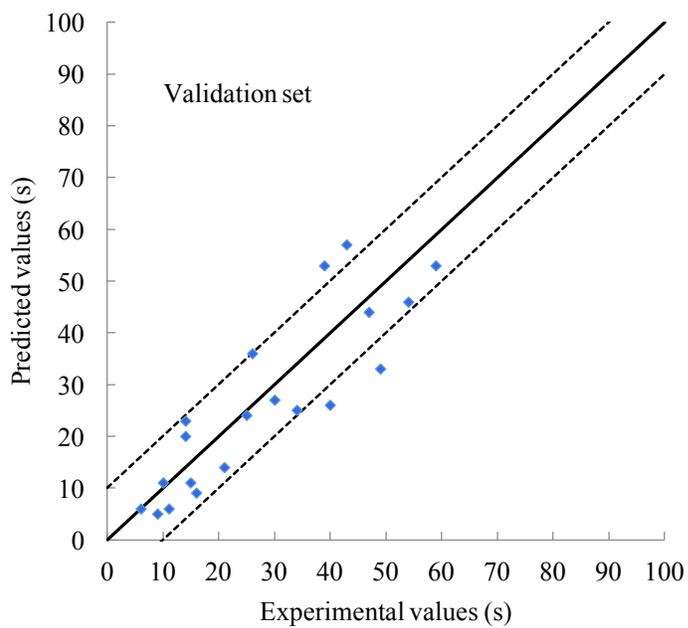

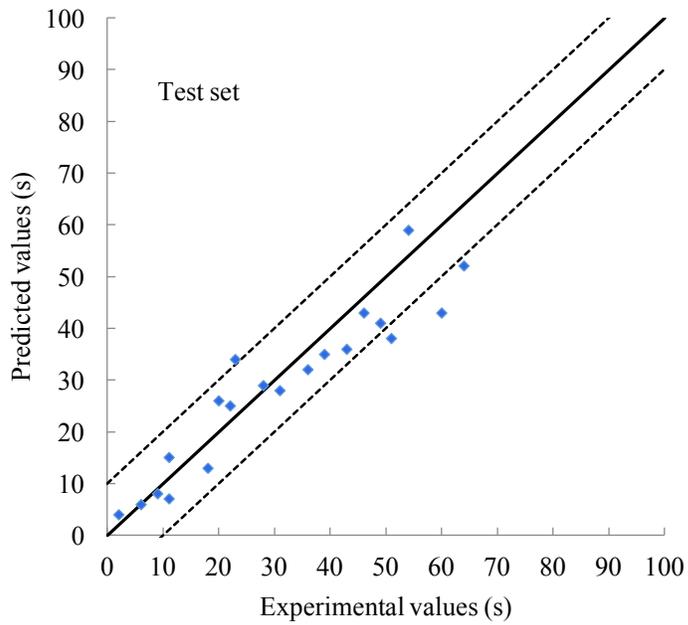

**Figure 3** Relationship between the experimental and the deep learning predicted values of the disintegration time on the OFDF training, validation and test sets. The dotted line indicates experimental values ± 10s.

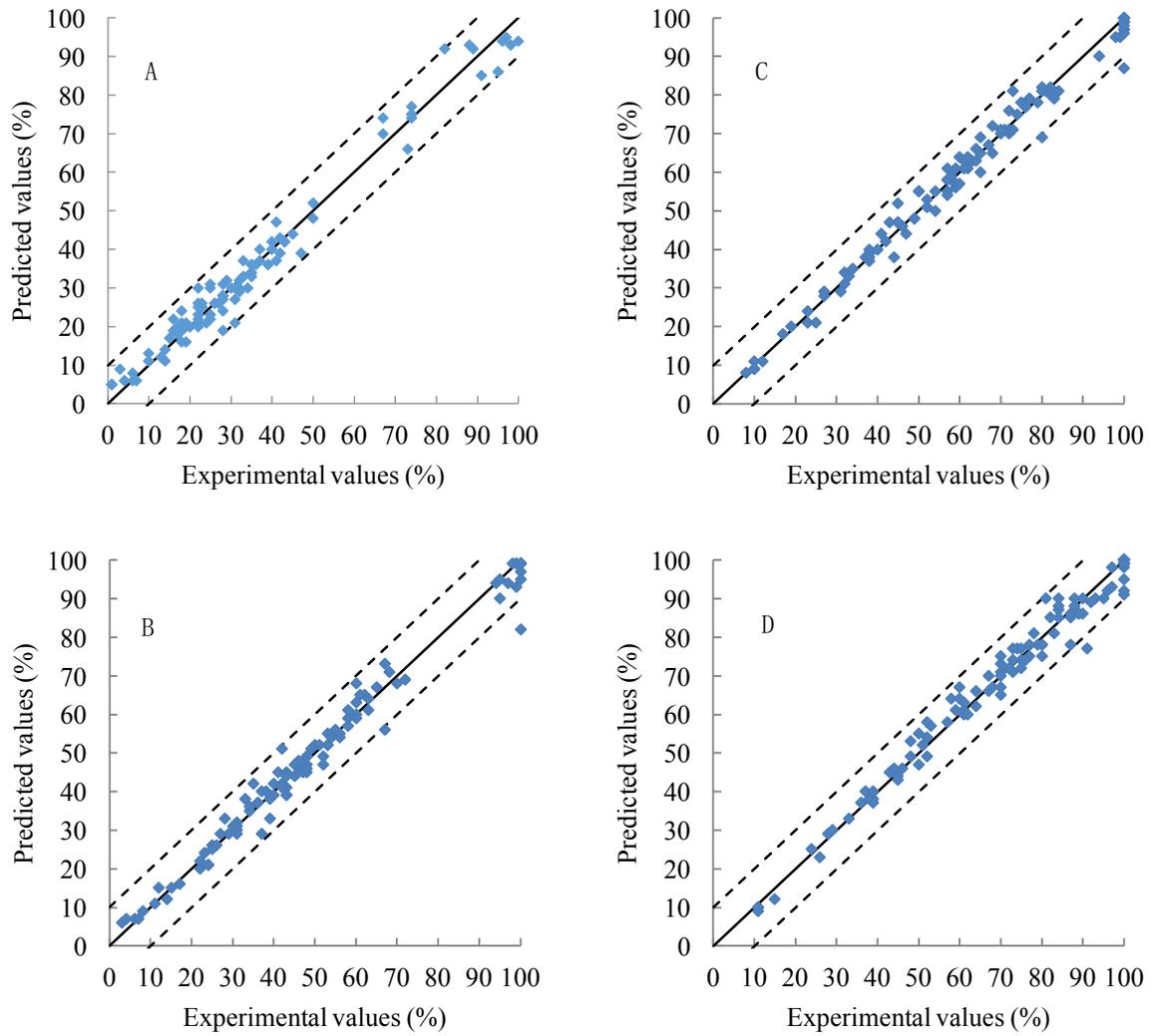

**Figure 4** Relationship between the experimental and the deep learning predicted values of the cumulative drug release percentages at 2h, 4h, 6h, 8h on the SRMT training set. A is for the values at 2h, B is for the values at 4h, C is for the values at 6h, D is for the values at 8h.

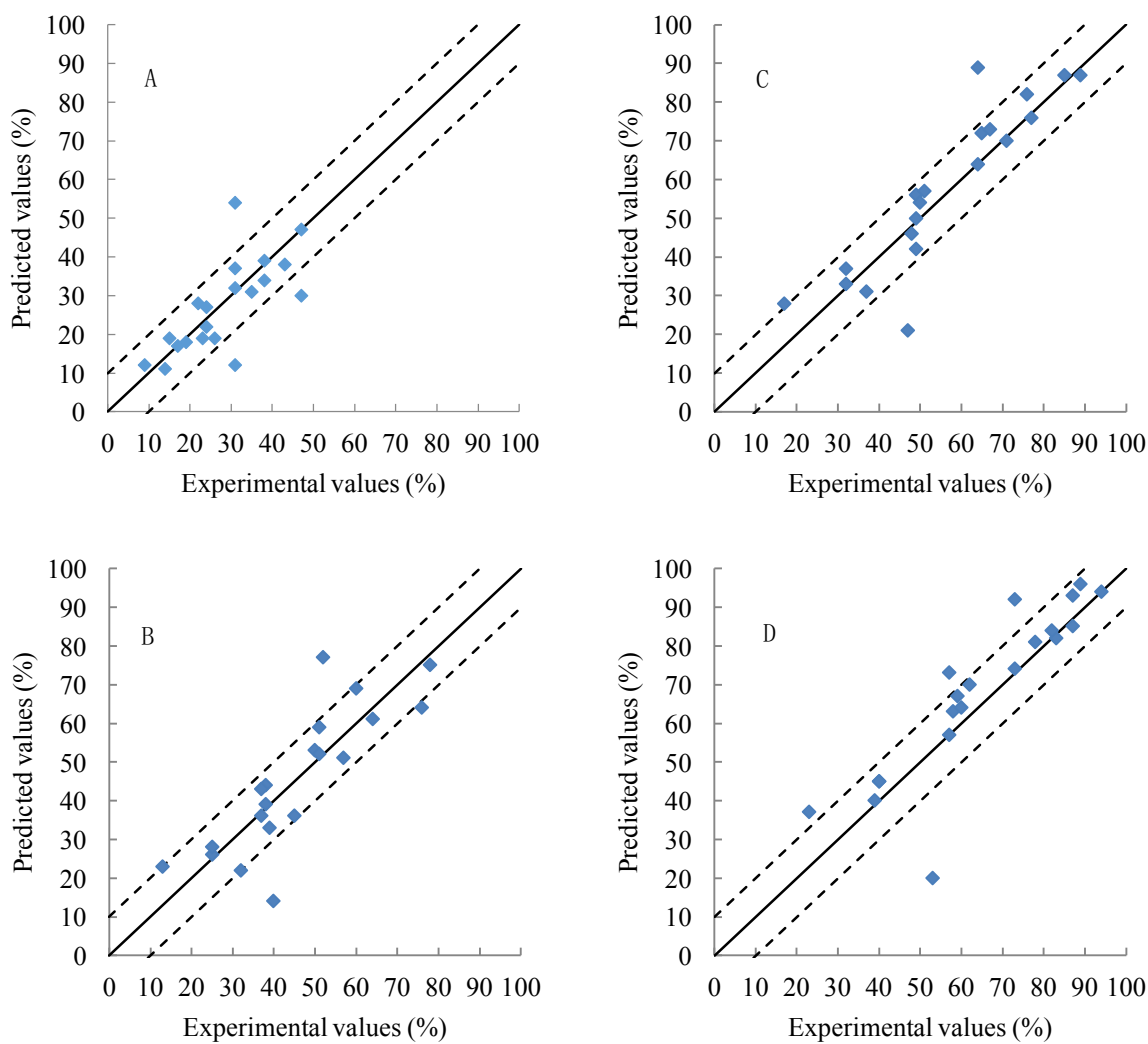

**Figure 5** Relationship between the experimental and the deep learning predicted values of the cumulative drug release percentages at 2h, 4h, 6h, 8h on the SRMT validation set. A is for the values at 2h, B is for the values at 4h, C is for the values at 6h, D is for the values at 8h.

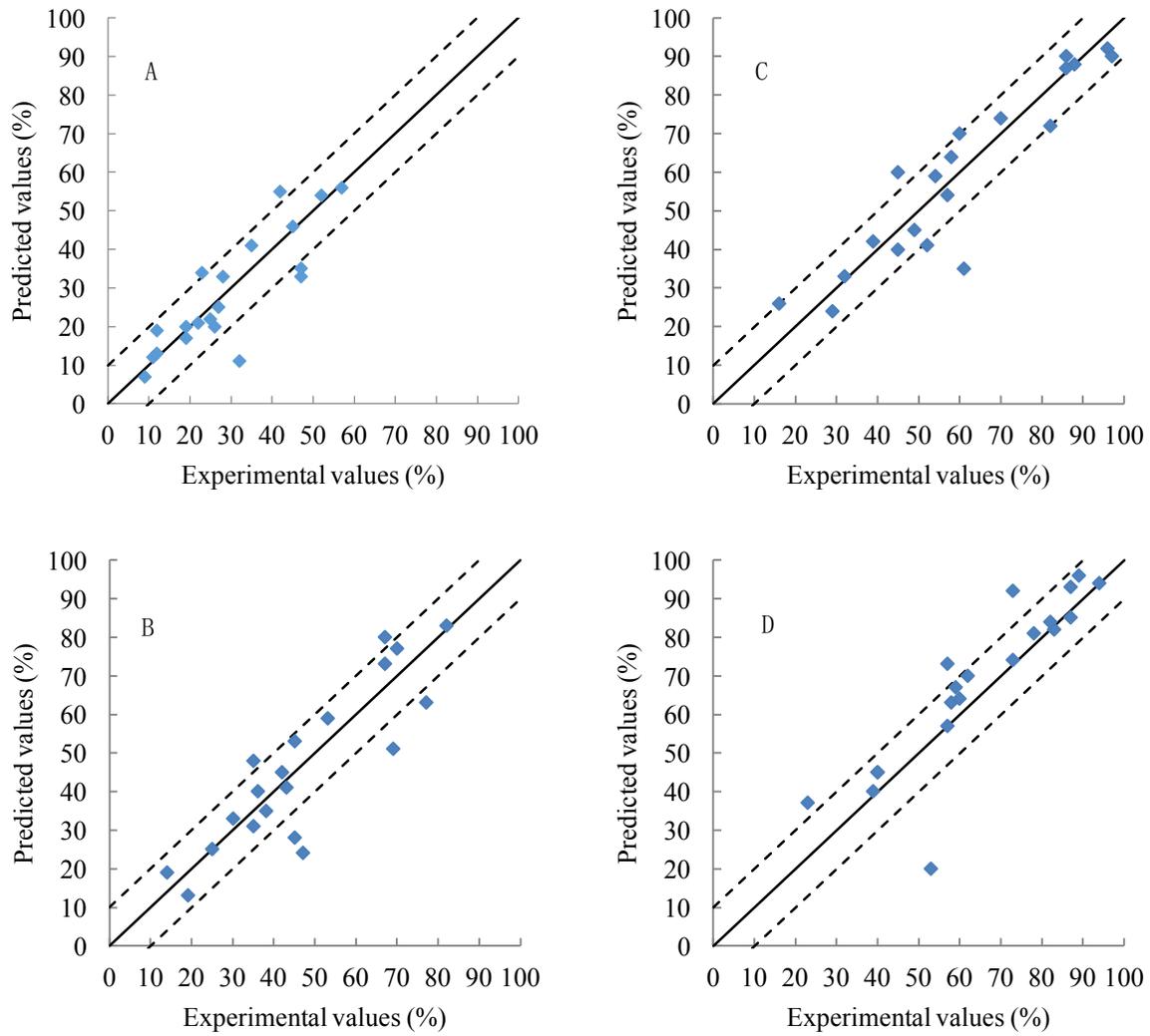

**Figure 6** Relationship between the experimental and the deep learning predicted values of the cumulative drug release percentages at 2h, 4h, 6h, 8h on the SRMT test set. A is for the values at 2h, B is for the values at 4h, C is for the values at 6h, D is for the values at 8h.

Multiple linear regression attempts to learn a linear combination of input features to predict the output. Multiple linear regression is simple and easy to model. The weights and biases of multiple linear regression could be directly calculated by using the least squares method. Multiple linear regression have better interpretability than other nonlinear machine learning models because the weights could indicate the importance of the input features in the prediction. However, multiple linear regression and partial least squared regression could only could only fit the linear function mapping, while obviously the relationship between the formulation and the key in vitro characteristics is complex and non-linear.

Random forest is an ensemble learning method. Ensemble learning methods combining multiple base learners could obtain better generalization ability of models than a single base learner. Random forest usually shows better performance than other ensemble learning models in many learning tasks. In random forest, the diversity of the base learners is not only from the sample disturbance but also from the attribute disturbance, which makes the difference between the base learners increase and the generalization ability be further improved. Support vector machine maps the sample from the original space to the higher dimensional feature space, therefore, the sample can be divided in the higher dimensional feature space. However, the conventional machine learning methods highly rely on the feature extractors designed by the subjective expert experiences.

Furthermore, the representative abilities of artificial neural networks enhance with the increase of the hidden layers and hidden nodes. The larger the model capacity, the more complex function the model can achieve. Therefore, deep learning containing more hidden layers could make multiple abstractions and feature extractions, which makes deep learning be able to accomplish more complex tasks to higher accuracy than the shallow artificial neural networks.

*3.6. Deep Learning in Formulation Prediction*

One main difficulty of formulation prediction is the lack of reliable and standard formulation data. The long experimental cycle and high cost of formulation development results in the small data set in this area. Moreover, current formulation researches focus on small amount of model drugs, which lead to highly imbalanced data space and further rise the difficulty of formulation prediction for other drugs. This fundamental issue was reported in previous research, in which the data sets are too small or too noisy.[30] To solve the issue, 10-fold cross validation was used for assessing the performance of the algorithm, which the $R^2$ value only reached 0.67 or 0.69.[30, 44] The prediction is even weaker in smaller data set because the small data set easily results in overfitting and poor generalizations.[30] Previous suggestion was to increase the amount of data set [30], but there are very limited formulation data due to the experimental limitation. Euclidean distance was used to estimate the domain of applicability of the trained model.[30] Another study indicated that maximum dissimilarity algorithm was able to select representative testing data of compounds from chemical databases.[42] However, both methods were good for the drug molecules, not for highly complexity formulation data. Deep learning with usual data selection algorithms and evaluation criteria is difficult to well predict with the small amount of formulation data with imbalanced input space. Therefore, MD-FIS algorithm is suitable for splitting formulation data with small sample size and imbalanced input space. In addition, two common evaluation metrics for regression problems (e.g. the correlation coefficient and coefficient of determination) cannot reflect the performance of pharmaceutical prediction models. Specific criteria need to be introduced for evaluating the model performance.

Currently there are only two deep learning researches in formulation prediction.[25, 44] In Zawbaa's research, 68 poly-lactide-co-glycolide formulations were used. Input vector contains 320 features and 745 release data at specific time points. At the beginning, the feature selection models were developed to minimize the input variables.[44] After that, seven machine learning approaches were compared, such as cubist, RF, ANNs, multivariate adaptive regression spline, classification and regression tree, and hybrid systems of fuzzy logic and evolutionary computations. The final results showed RF had the best performance with 0.692 coefficient of determination. Moreover, the prediction of the model was only suitable for 2/3 formulations and the rest 1/3 formulations had high error. The dissolution profiles of 4 proteins among total 14 proteins were strongly not recommended by the final model. Formulation development contains high dimensional factors, such as drug diversity, excipient types, drug/excipient ratios, dosage forms, manufacturing processes and multiple characteristics, which lead to the high complexity data. Actually, deep learning can automatically learn high level features from data without explicitly providing feature selection model. Combining feature extractors with deep learning model is unnecessary and inappropriate. Therefore, this research still needs significant improvement on methodology for formulation prediction. Recently deep learning with MD-FIS was applied for formulation prediction of oral fast disintegrating tablets.[25] The results showed that the prediction of deep learning was better than that of ANN because deep learning could extract the features automatically without designing feature extractors. In addition, there are some researches about machine learning methods in formulation design, as summarized in Table 1. For example, an expert system was constructed to predict formulations of oral disintegrating tablets with 15 input parameters.[12, 13] Zhi-hong Zhang et al. built an expert system with ANN for the formulation development of oral osmotic pump controlled release tablets. [5, 14] QSAR models were developed to predict the binding affinity of β-cyclodextrin and sulfobutylether-β-cyclodextrin complexation. However, these results indicated that the conventional machine learning methods need feature extraction before the prediction models and didn't showed good prediction performance.

## 4. Conclusion

In this paper, deep learning models were successfully developed to predict pharmaceutical formulations on small data. The good generalization performance of the models was demonstrated by the external datasets. The proposed models could effectively predict the key characteristics in regression problems than the models trained by other machine learning methods, because deep learning can find out the complex correlation in the data. Quality by design (QbD) is one concept that the idea of the pharmaceutical development should be paid attention throughout the drug development process. Machine learning methods could not only help to predict the in vivo and in vitro characteristics based on the formulation and process data, but also assist in the pharmaceutical experiment design and help to control the product quality in the whole product cycle. Deep learning shows a broad prospect in the implementation of QbD. We expect deep learning could significantly shorten the drug product development timeline and decrease the material usage. Furthermore, the cross-disciplinary integration of pharmaceutics and artificial intelligence may shift the paradigm of pharmaceutical researches from experience-dependent studies to data-driven methodologies. In the future, our laboratory would investigate other machine learning methods (e.g. transfer learning) for formulation prediction to achieve better performance.

## Acknowledgments

Current research is financially supported by the University of Macau Research Grant (MYRG2016-00038-ICMS-QRCM, MYRG2016-00040-ICMS-QRCM and MYRG2017-00141-FST), Macau Science and Technology Development Fund (FDCT) (Grant No. 103/2015/A3) and the National Natural Science Foundation of China (Grant No. 61562011).

## References

1    Zhang W, Zhao Q, Deng J, Hu Y, Wang Y, Ouyang D. Big data analysis of global advances in pharmaceutics and drug delivery 1980–2014. *Drug Discov Today* 2017;**22**:1201-8.
2    Seddon G, Lounnas V, McGuire R, Van Den Bergh T, Bywater RP, Oliveira L, *et al*. Drug design for ever, from hype to hope. *J. Comput-Aided Mol Des* 2012;**26**:137-50.
3    Ouyang D, Smith SC. Computational pharmaceutics: Application of molecular modeling in drug delivery. Computational pharmaceutics: Application of molecular modeling in drug delivery. wiley; 2015, p. 1-303.
4    Rowe RC, Roberts RJ. Artificial intelligence in pharmaceutical product formulation: Knowledge-based and expert systems. *Pharm Sci Technol Today* 1998;**1**:153-9.
5    Zhang ZH, Pan WS. Expert system for the development and formulation of push-pull osmotic pump tablets containing poorly water-soluble drugs. In: Formulation tools for pharmaceutical development. Elsevier Ltd; 2013, p. 73-108.
6    Mendyk A, Szlęk J, Jachowicz R. Me_expert 2.0: A heuristic decision support system for microemulsions formulation development. In: Formulation tools for pharmaceutical development. Elsevier Ltd; 2013, p. 39-71.
7    Han X, Jiang H, Han L, Xiong X, He Y, Fu C, *et al*. A novel quantified bitterness evaluation model for traditional chinese herbs based on an animal ethology principle. *Acta pharmaceutica sinica B* 2018;**8**:209-17.
8    Wang Z, He Z, Zhang L, Zhang H, Zhang M, Wen X, *et al*. Optimization of a doxycycline hydroxypropyl-β-cyclodextrin inclusion complex based on computational modeling. *Acta Pharmaceutica Sinica B* 2013;**3**:130-9.
9    Xu W-J, Li N, Gao C-k. Preparation of controlled porosity osmotic pump tablets for salvianolic acid and optimization of the formulation using an artificial neural network method. *Acta Pharmaceutica Sinica B* 2011;**1**:64-70.
10   Suñé Neǵre JM, Roiǵ Carreras M, García RF, Montoya EG, Lozano PP, Aǵuilar JE, *et al*. Sedem diagram: An expert system for preformation, characterization and optimization of tablets obtained by direct compression. In: Formulation tools for pharmaceutical development. Elsevier Ltd; 2013, p. 109-35.
11   Wilson WI, Peng Y, Augsburger LL. Generalization of a prototype intelligent hybrid system for hard gelatin capsule formulation development. *AAPS PharmSciTech [electronic resource]*. 2005;**6**:E449-57.
12   Aguilar-Díaz JE, García-Montoya E, Suñe-Negre JM, Pérez-Lozano P, Miñarro M, Ticó JR. Predicting orally disintegrating tablets formulations of ibuprophen tablets: An application of the new sedem-odt expert system. *Eur J Pharm Biopharm* 2012;**80**:638-48.
13   Aguilar JE, Montoya EG, Lozano PP, Negre JMS, Carmona MM, Grau JRT. New sedem-odt expert system: An expert system for formulation of orodispersible tablets obtained by direct compression. In: Formulation tools for pharmaceutical development. Elsevier Ltd; 2013, p. 137-54.
14   Zhang ZH, Dong HY, Peng B, Liu HF, Li CL, Liang M, *et al*. Design of an expert system for the development and formulation of push-pull osmotic pump tablets containing poorly water-soluble drugs. *Int J Pharm* 2011;**410**:41-7.
15   Chalortham N, Ruangrajitpakorn T, Supnithi T, Leesawat P. Oxpirt: Ontology-based expert system for production of a generic immediate release tablet. In: Formulation tools for pharmaceutical development. Elsevier Ltd; 2013, p. 203-28.
16   Trnka H, Wu JX, Van De Weert M, Grohganz H, Rantanen J. Fuzzy logic-based expert system for evaluating cake quality of freeze-dried formulations. *J Pharm Sci* 2013;**102**:4364-74.

17      Merzlikine A, Abramov YA, Kowsz SJ, Thomas VH, Mano T. Development of machine learning models of $\beta$-cyclodextrin and sulfobutylether-β-cyclodextrin complexation free energies. *Int J Pharm* 2011;**418**:207-16.
18      Bengio Y, Courville A, Vincent P. Representation learning: A review and new perspectives. *IEEE Trans Pattern Anal Mach Intell* 2013;**35**:1798-828.
19      Lecun Y, Bengio Y, Hinton G. Deep learning. *Nature* 2015;**521**:436-44.
20      Schmidhuber J. Deep learning in neural networks: An overview. *Neural Netw* 2015;**61**:85-117.
21      Krizhevsky A, Sutskever I, Hinton GE. p 1097-105.
22      Hinton G, Deng L, Yu D, Dahl G, Mohamed AR, Jaitly N*, et al*. Deep neural networks for acoustic modeling in speech recognition: The shared views of four research groups. *IEEE Signal Process Mag* 2012;**29**:82-97.
23      Bengio Y, Simard P, Frasconi P. Learning long-term dependencies with gradient descent is difficult. *IEEE Trans. Neural Networks* 1994;**5**:157-66.
24      Sutskever I, Vinyals O, Le QV. In: Ghahramani Z, Weinberger KQ, Cortes C, Lawrence ND, Welling M, Editors. (Neural information processing systems foundation, January ed), p. 3104-12.
25      Han R, Yang Y, Li X, Ouyang D. Predicting oral disintegrating tablet formulations by neural network techniques. *Asian Journal of Pharmaceutical Sciences* 2018.
26      Ekins S. The next era: Deep learning in pharmaceutical research. *Pharm. Res.* 2016;**33**:2594-603.
27      Mamoshina P, Vieira A, Putin E, Zhavoronkov A. Applications of deep learning in biomedicine. *Mol Pharm* 2016;**13**:1445-54.
28      Baskin II, Winkler D, Tetko IV. A renaissance of neural networks in drug discovery. *Expert Opin. Drug Discov* 2016;**11**:785-95.
29      Gawehn E, Hiss JA, Schneider G. Deep learning in drug discovery. *Mol Inform* 2016;**35**:3-14.
30      Lusci A, Pollastri G, Baldi P. Deep architectures and deep learning in chemoinformatics: The prediction of aqueous solubility for drug-like molecules. *J Chem Inf Model* 2013;**53**:1563-75.
31      Hughes TB, Miller GP, Swamidass SJ. Modeling epoxidation of drug-like molecules with a deep machine learning network. *ACS Cent Sci* 2015;**1**:168-80.
32      Xu Y, Dai Z, Chen F, Gao S, Pei J, Lai L. Deep learning for drug-induced liver injury. *J Chem Inf Model* 2015;**55**:2085-93.
33      Ma J, Sheridan RP, Liaw A, Dahl GE, Svetnik V. Deep neural nets as a method for quantitative structure-activity relationships. *J Chem Inf Model* 2015;**55**:263-74.
34      Ramsundar B, Liu B, Wu Z, Verras A, Tudor M, Sheridan RP*, et al*. Is multitask deep learning practical for pharma? *J Chem Inf Model* 2017;**57**:2068-76.
35      Altae-Tran H, Ramsundar B, Pappu AS, Pande V. Low data drug discovery with one-shot learning. *ACS Cent Sci* 2017;**3**:283-93.
36      Aliper A, Plis S, Artemov A, Ulloa A, Mamoshina P, Zhavoronkov A. Deep learning applications for predicting pharmacological properties of drugs and drug repurposing using transcriptomic data. *Mol Pharm* 2016;**13**:2524-30.
37      Vanhaelen Q, Mamoshina P, Aliper AM, Artemov A, Lezhnina K, Ozerov I*, et al*. Design of efficient computational workflows for in silico drug repurposing. *Drug Discov Today* 2017;**22**:210-22.
38      Korotcov A, Tkachenko V, Russo DP, Ekins S. Comparison of deep learning with multiple machine learning methods and metrics using diverse drug discovery data sets. *Mol Pharm* 2017;**14**:4462-75.
39      Pedregosa F, Varoquaux G, Gramfort A, Michel V, Thirion B, Grisel O*, et al*. Scikit-learn: Machine learning in python. *J Mach Learn Res* 2011;**12**:2825-30.
40      Shah VP, Tsong Y, Sathe P, Liu JP. In vitro dissolution profile comparison- statistics and analysis of the similarity factor, f2. *Pharm Res* 1998;**15**:889-96.

41	Martin TM, Harten P, Young DM, Muratov EN, Golbraikh A, Zhu H, *et al*. Does rational selection of training and test sets improve the outcome of qsar modeling? *J Chem Inf Model* 2012;**52**:2570-8.
42	Snarey M, Terrett NK, Willett P, Wilton DJ. Comparison of algorithms for dissimilarity-based compound selection. *J Mol Graph Model* 1997;**15**:372-85.
43	Willett P. Dissimilarity-based algorithms for selecting structurally diverse sets of compounds. *J Comput Biol* 1999;**6**:447-57.
44	Zawbaa HM, Szlęk J, Grosan C, Jachowicz R, Mendyk A. Computational intelligence modeling of the macromolecules release from plga microspheres-focus on feature selection. *PLoS One* 2016;**11**.